\documentclass[final]{nesy2026} 

\usepackage{tikz}
\usepackage{booktabs}
\usepackage{multirow}
\usetikzlibrary{arrows.meta, positioning, shapes.geometric, calc, fit, backgrounds}
\SetCommentSty{textrm}  
\SetKwComment{tcp}{$\triangleright$}{}  

\usepackage{longtable}
\usepackage{placeins} 

\usepackage{siunitx}

\theorembodyfont{\upshape}
\theoremheaderfont{\scshape}
\theorempostheader{:}
\theoremsep{\newline}

\title[Constraint-Guided Enterprise Data Mapping with LLMs]{Constraint-Guided Enterprise Data Mapping\\with Large Language Models}

\ifanonsubmission
\else
\author{\Name{Sebastian Monka\nametag{$^{1}$}} \Email{sebastian.monka@de.bosch.com}\\
  \Name{Pramod Anantharam\nametag{$^{2}$}} \Email{pramod.anantharam@us.bosch.com}\\
  \Name{Thien {Vo Minh}\nametag{$^{3}$}} \Email{Thien.VoMinh@vn.bosch.com}\\
  \Name{Lavdim Halilaj\nametag{$^{1}$}} \Email{lavdim.halilaj@de.bosch.com}\\
  \addr $^{1}$Bosch Center for Artificial Intelligence, Renningen, Germany \quad
  $^{2}$Bosch Center for Artificial Intelligence, Pittsburgh, USA \quad
  $^{3}$Robert Bosch GmbH, Ho Chi Minh City, Vietnam}
\fi

\begin{document}

\maketitle

\begin{abstract}
  Enterprise entity alignment must handle semi-structured records, implicit attributes, and unit or granularity mismatches. Manual matching is still common in practice, but it does not scale as schemas and providers evolve. LLM-only matching improves semantic recall, yet it can violate structural and physical invariants, producing fluent yet operationally invalid correspondences.

  We propose \emph{constraint-guided mapping} (CGM), a neuro-symbolic method with three stages: (i) schema-grounded admissibility constraints with metadata $m_c=\langle\tau_c,\delta_c\rangle$, where $\tau_c$ denotes the constraint type and $\delta_c$ provides executable relation and normalization logic; (ii) constraint-restricted candidate generation with cascade relaxation to guarantee a nonempty feasible set under noise; and (iii) neural ranking with bounded LLM disambiguation restricted to that feasible set.

  Methodologically, constraints operate as hypothesis-space operators rather than post-hoc validators, enabling controlled degradation under relaxation and auditable, human-guidable decisions. On a controlled structural-decoy benchmark, hard admissibility shrinks the candidate space ${\sim}480\times$ without dropping the GT, and a layer-by-layer ablation shows this gate---not the LLM---is the decisive lift (F1 $0.08\!\to\!0.66$). The benefit is model-independent and adds no extra inference cost: a small model with constraints matches a frontier LLM used \emph{without} them at ${\sim}28\times$ lower cost. The method---not a single tuned configuration---transfers across seven enterprise makes (macro F1 $0.70$), each under its own automatically discovered, expert-refinable constraints, and lowers expert effort by ${\sim}7\times$ versus spreadsheet workflows. Public Valentine results add an external ranking sanity check and mark the boundary: constraints should be hard only where structural invariants are match-determining.
\end{abstract}

\section{Introduction}

Enterprise data products must align heterogeneous schemas from different organizations, legacy systems, and evolving conventions. Although schema matching and entity resolution are well studied \cite{Rahm2001ASO, Bellahsene2013SchemaMA}, enterprise alignment remains largely manual as schemas, providers, and naming conventions continuously change.

Prior methods combine schema- and instance-level signals, often as independently scored components; we instead formalize enterprise alignment as constraint-guided correspondence inference, where symbolic constraints define admissible candidates and neural components rank and disambiguate only within this bounded hypothesis space.

As a motivating example, aligning a source record \texttt{Panda 141 i.e.} (\texttt{engine\_cc}=$1000$, year range \texttt{1991--1996}) to the target \texttt{Fiat Panda 141} (\texttt{engine\_size}=\texttt{1.0\,L}, build date \texttt{20.08.1993}) requires implicit-attribute extraction, unit conversion ($1000\,$cc${=}1.0\,$L), and granularity matching---not lexical similarity. Such failure modes defeat lexical or unconstrained neural matching. We address them with CGM: executable symbolic constraints ($\delta_c$) define the admissible space, and neural components rank and disambiguate only within it.

\paragraph{Contributions.}
(1)~A constraint-\emph{preconditioned} neuro-symbolic method: hard constraints act as hypothesis-space operators \emph{before} neural reasoning, with cascade relaxation guaranteeing a nonempty feasible set and bounded LLM disambiguation valid \emph{by construction}.
(2)~A controlled study of \emph{why} and \emph{when} constraints help: competitive with SOTA matchers on public schema matching, and---via a released synthetic benchmark, a layer-by-layer ablation, and an ambiguity-trap analysis---showing that where numbers, units, and identifiers decide the match, probabilistic matching is misled while admissibility stays valid by construction, with the hard gate the decisive layer (F1 $0.08\!\to\!0.66$).
(3)~A model ablation showing the benefit is model-independent and adds \emph{no extra per-mapping LLM calls}: a small model with constraints reaches the same $100\%$ validity as a frontier model at $\sim$$28\times$ lower cost.
(4)~An enterprise deployment reducing expert effort by $\sim$$7\times$, with implementation and benchmark released.

\section{Problem Setting and Requirements}
\label{sec:problem-setting}

We align source entities to canonical targets under schemas $S_s,S_t$. Compared to curated benchmarks, enterprise data couples three heterogeneity classes: \emph{representation} (composite/implicit fields), \emph{structure} (unit and aggregation mismatch), and \emph{semantics} (abbreviations, multilingual variants, temporal drift)---the failure modes of the motivating example above. Representation and structure define admissibility; semantics is resolved within admissible candidates.

These demand invariants beyond lexical similarity, yielding four requirements: (1)~\emph{controlled semantic decomposition} of implicitly encoded fields; (2)~\emph{explicit constraint enforcement} for unit and granularity compatibility; (3)~\emph{graceful degradation} when constraints are incomplete; and (4)~\emph{auditability} of the contributing constraints and signals. Together these motivate a constraint-guided pipeline with hard admissibility, cascade relaxation, and bounded neural disambiguation.

\section{Related Work}

Entity matching and alignment are long-standing data-integration problems \cite{Rahm2001ASO, Bellahsene2013SchemaMA}; Valentine provides a benchmarking framework \cite{Koutras2021Valentine} and recent surveys summarize LLM opportunities and limits \cite{Freire2025Survey}. The distinguishing axis across paradigms is \emph{when} structural validity is enforced.

\emph{Constraint/rule systems} enforce domain invariants through declarative specifications: reliable when assumptions hold, but brittle under paraphrases, sparse metadata, and heterogeneous naming \cite{Qi2025CleanAgent, Khoee2025GateLens}. \emph{PLM- and LLM-based matchers} improve semantic flexibility (Ditto and follow-ups \cite{Li2020Ditto, Doehmen2024SchemaPile, Xu2024KcMF, Zhang2023NameGuess, Parciak2025LLMMatcher}; profile-based reasoning \cite{Peeters2023EntityMatchingLLM}) but typically score correspondences without a hard admissibility gate and may propagate representation/normalization errors. \emph{Constrained, grounded neuro-symbolic pipelines}---constrained decoding (CRANE) and retrieval-/knowledge-grounded methods (ReMatch, KG-RAG4SM, SMoG, MATP) \cite{Banerjee2025CRANE, Sheetrit2024ReMatch, Ma2025KGRAG4SM, Jeon2025SMoG, MATP2025}---improve reliability but operate at the token level or over a softly filtered pool. Magneto, the closest baseline in scope, targets scalable enterprise mapping via retrieval and iterative refinement \cite{Freire2025Magneto}.

Unlike all of these, we treat constraints as \emph{hypothesis-space operators applied before} neural reasoning---schema-grounded hard admissibility (with $\delta_c$-encoded normalization) over the entity-level candidate space, with cascade relaxation guaranteeing a nonempty feasible set and traceable degradation. For comparison we (i)~run SOTA Magneto \cite{Freire2025Magneto} and standard matchers (COMA, Cupid, SimilarityFlooding, Jaccard) on Valentine, and adapt the same Magneto to record-level matching on our synthetic and enterprise benchmarks (Appendix~\ref{app:hetero}); (ii)~on enterprise data use an unconstrained retrieval-augmented LLM matcher as a faithful \emph{proxy} (our \textbf{LLM} baseline: rule-informed retrieval $+$ LLM decision \emph{without} gating); and (iii)~release a synthetic benchmark (Section~\ref{sec:synthetic}) isolating the core heterogeneity classes.

\section{Problem Formalization}
\label{sec:problem-formalization}

We formalize enterprise alignment as constraint-guided correspondence inference. Let $E_s,E_t$ be source/target entity sets with schema attributes $S_s,S_t$. The ground-truth alignment is a set-valued map $M:E_s\to 2^{E_t}$, where $e_t\in M(e_s)$ iff $e_s,e_t$ are semantically and structurally consistent. The system returns a prediction $\tilde{M}(e_s)$ drawn from a bounded shortlist $\hat{M}(e_s)$ with $|\hat{M}(e_s)|=k\ll|E_t|$.

\paragraph{Schema-grounded constraints.}
Domain knowledge is a set of constraints $C$, each $c=\langle a_s^i,\, m_c,\, a_t^j\rangle$ referencing a source and target attribute with metadata $m_c=\langle\tau_c,\delta_c\rangle$: type $\tau_c\in\{\text{hard},\text{soft}\}$ and executable relation/normalization logic $\delta_c$. Instantiated from $\delta_c$ and representative examples (then expert-reviewed), each constraint is a predicate $c:(E_s(a_s^i),E_t(a_t^j))\to\{0,1\}$ enforcing compatibility (identifier, unit, aggregation). Hard constraints $\tilde{C}_h$ eliminate invalid correspondences; soft constraints $\tilde{C}_s$ contribute ranking evidence ($\tilde{\cdot}$ denotes the executable, instantiated form).

\paragraph{Admissible space and cascade relaxation.}
For $e_s$, the admissible space under hard constraints is $\mathcal{H}_{\tilde{C}_h}(e_s)=\{e_t\in E_t \mid \forall c\in\tilde{C}_h:\ c(e_s,e_t)=1\}$. To handle noise, we relax hard constraints in priority order, taking the largest subset $\tilde{C}_h^{*}$ with $\mathcal{H}_{\tilde{C}_h^{*}}(e_s)\neq\emptyset$ and setting $\mathcal{H}(e_s):=\mathcal{H}_{\tilde{C}_h^{*}}(e_s)$. Relaxation is monotone ($\tilde{C}_h'\subseteq\tilde{C}_h \Rightarrow \mathcal{H}_{\tilde{C}_h}(e_s)\subseteq\mathcal{H}_{\tilde{C}_h'}(e_s)$), so it terminates in at most $|\tilde{C}_h|$ steps; if all hard constraints are dropped ($\tilde{C}_h^{*}=\emptyset$) then $\mathcal{H}(e_s)=E_t$ and the method reduces to unconstrained retrieval.

\paragraph{Prediction and evaluation.}
The system predicts $\tilde{M}(e_s)\subseteq\hat{M}(e_s)\subseteq\mathcal{H}(e_s)$: hard constraints (with relaxation) induce $\mathcal{H}$, neural ranking forms the shortlist $\hat{M}$, and disambiguation resolves $\tilde{M}$ within it. A prediction is \emph{complete} ($\tilde{M}(e_s)=M(e_s)$), \emph{partial} ($M(e_s)\subset\tilde{M}(e_s)$), or \emph{wrong} otherwise.

\section{Approach}
\label{sec:approach}

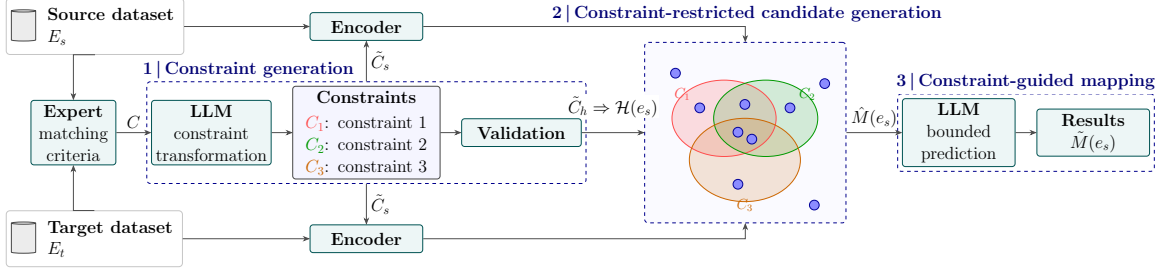
\begin{figure}[t]
  \centering
  \resizebox{\linewidth}{!}{%
    \begin{tikzpicture}[
        font=\Large,
        >=Stealth,
        node distance=6mm and 10mm,
        line/.style={->, line width=0.5pt, draw=gray!60!black},
        block/.style={draw=gray!55!black, fill=blue!4, rounded corners=2.5pt, align=center, inner sep=3.5pt, minimum height=8mm, text width=30mm},
        blockWide/.style={block, text width=40mm},
        dsbox/.style={draw=gray!45, rounded corners=3pt, inner sep=4pt},
        dslabel/.style={align=left},
        db/.style={cylinder, shape border rotate=90, aspect=0.25, draw=gray!55!black, fill=gray!15, minimum height=9mm, minimum width=7mm},
        accent/.style={draw=teal!65!black, fill=teal!7, line width=0.8pt},
        dashedbox/.style={draw=blue!55!black, dashed, line width=0.7pt, rounded corners=3pt, inner sep=6pt},
        grouptitle/.style={font=\Large\bfseries, text=blue!45!black, fill=white, inner sep=2pt},
        dot/.style={circle, draw=blue!60!black, fill=blue!45, inner sep=0pt, minimum size=2.8mm}
      ]

      \node[db] (srcdb) {};
      \node[dslabel, right=2mm of srcdb] (srctxt) {\textbf{Source dataset}\\$E_s$};
      \node[dsbox, fit=(srcdb)(srctxt)] (src) {};

      \node[db, below=52mm of srcdb] (tgtdb) {};
      \node[dslabel, right=2mm of tgtdb] (tgttxt) {\textbf{Target dataset}\\$E_t$};
      \node[dsbox, fit=(tgtdb)(tgttxt)] (tgt) {};

      \node[block, accent, right=2mm of $(srcdb)!0.5!(tgtdb)$, text width=22mm] (expnode) {\textbf{Expert}\\matching criteria};
      \node[block, accent, right=10mm of expnode, text width=32mm] (rgllm) {\textbf{LLM}\\constraint\\transformation};
      \node[blockWide, right=6mm of rgllm] (constraints) {\textbf{Constraints}\\[1mm]
        {\textcolor{red!70}{$C_1$}: constraint 1}\\
        {\textcolor{green!60!black}{$C_2$}: constraint 2}\\
      {\textcolor{orange!80!black}{$C_3$}: constraint 3}};
      \node[block, accent, right=6mm of constraints, text width=32mm] (validate) {\textbf{Validation}};

      \node[dashedbox, fit=(rgllm)(constraints)(validate), inner sep=3.5pt,
      label={[grouptitle, xshift=-34mm]north:\textbf{1\,\textbar\,Constraint generation}}] (rgbox) {};

      \node[block, accent] (embS) at (src.center -| constraints.center) {\textbf{Encoder}};
      \node[block, accent] (embT) at (tgt.center -| constraints.center) {\textbf{Encoder}};

      \node[dashedbox, fill=blue!2, right=18mm of validate, minimum width=58mm, minimum height=52mm, anchor=west, label={[grouptitle, xshift=0.5mm, yshift=5mm]north:\textbf{2\,\textbar\,Constraint-restricted candidate generation}}] (space) {};

      \path (space.north west) ++(23mm,-22mm) coordinate (c1);
      \path (space.north west) ++(35mm,-22mm) coordinate (c2);
      \path (space.north west) ++(29mm,-34mm) coordinate (c3);

      \fill[red!70, opacity=0.18]   (c1) ellipse (15mm and 11mm);
      \fill[green!60!black, opacity=0.16] (c2) ellipse (15mm and 11mm);
      \fill[orange!80!black, opacity=0.16] (c3) ellipse (16mm and 12mm);

      \draw[red!70, line width=0.55pt]   (c1) ellipse (15mm and 11mm);
      \draw[green!60!black, line width=0.55pt] (c2) ellipse (15mm and 11mm);
      \draw[orange!80!black, line width=0.55pt] (c3) ellipse (16mm and 12mm);

      \node[font=\large, text=red!70] at ($(space.north west)+(11mm,-15mm)$) {$C_1$};
      \node[font=\large, text=green!60!black] at ($(space.north west)+(47mm,-15mm)$) {$C_2$};
      \node[font=\large, text=orange!80!black] at ($(space.north west)+(29mm,-47mm)$) {$C_3$};

      \node[dot] at ($(space.north west)+(9mm,-9mm)$) {};
      \node[dot] at ($(space.north west)+(52mm,-12mm)$) {};
      \node[dot] at ($(space.north west)+(49mm,-47mm)$) {};
      \node[dot] at ($(space.north west)+(16mm,-19mm)$) {};
      \node[dot] at ($(space.north west)+(42mm,-19mm)$) {};
      \node[dot] at ($(space.north west)+(27mm,-41mm)$) {};
      \node[dot] at ($(space.north west)+(29mm,-18mm)$) {};
      \node[dot] at ($(space.north west)+(27mm,-26mm)$) {};
      \node[dot] at ($(space.north west)+(31mm,-28mm)$) {};

      \node[block, right=16mm of space, accent] (llm) {\textbf{LLM}\\bounded prediction};
      \node[block, right=6mm of llm, accent] (map) {\textbf{Results}\\$\tilde{M}(e_s)$};

      \node[dashedbox, fit=(llm)(map), inner sep=3.5pt, label={[grouptitle, xshift=0mm]north:\textbf{3\,\textbar\,Constraint-guided mapping}}] (predbox) {};

      \draw[line] (src.south) -| (expnode.north);
      \draw[line] (tgt.north) -| (expnode.south);
      \draw[line] (expnode.east) -- node[midway, above] {$C$}
      (rgllm.west);
      \draw[line] (rgllm.east) -- (constraints.west);
      \draw[line] (constraints.east) -- (validate.west);

      \draw[line] (src.east) -- (embS.west);
      \draw[line] (tgt.east) -- (embT.west);
      \draw[line] (rgbox.north) -- node[midway, right] {$\tilde{C}_s$}
      (embS.south);
      \draw[line] (rgbox.south) -- node[midway, right] {$\tilde{C}_s$}
      (embT.north);

      \draw[line] (embS.east) -| (space.north);
      \draw[line] (embT.east) -| (space.south);
      \draw[line] (validate.east) -- node[midway, above, fill=white, inner sep=1pt, yshift=4mm] {$\tilde{C}_h\Rightarrow\mathcal{H}(e_s)$}
      (space.west);

      \draw[line] (space.east) -- node[midway, above] {$\hat{M}(e_s)$}
      (llm.west);
      \draw[line] (llm.east) -- (map.west);

    \end{tikzpicture}%
  }%
  \caption{CGM as a constrained funnel: hard constraints $\tilde{C}_h$ shrink the catalog to the admissible set $\mathcal{H}(e_s)$, soft-constraint neural ranking forms the shortlist $\hat{M}(e_s)$, and the LLM selects only within it---so $\tilde{M}(e_s)\subseteq\hat{M}(e_s)\subseteq\mathcal{H}(e_s)$ by construction.}
  \label{fig:approach-overview}
\end{figure}

CGM applies a constraint-guided design in which executable hard constraints $\tilde{C}_h$ define $\mathcal{H}(e_s)$ before neural ranking and LLM disambiguation, so neural inference runs \emph{under} symbolic admissibility ($\tilde{M}(e_s)\subseteq\hat{M}(e_s)\subseteq\mathcal{H}(e_s)$)---preserving semantic flexibility while guaranteeing structural consistency. Figure~\ref{fig:approach-overview} overviews the pipeline.

\subsection{Constraint Definition and Validation}
\label{sec:constraint-generation}

Expert knowledge is encoded as executable constraints that specify the referenced source and target attributes, a transformation or constraint expression, and a type $\tau \in \{\text{hard},\text{soft}\}$. Hard constraints define admissibility; soft constraints provide ranking signals (Appendix~\ref{app:synthetic-example} works one hard and one soft rule end to end, Table~\ref{tab:constraint-mechanics}).

\paragraph{Expert-assisted constraint specification.}
The expert specifies a constraint skeleton by selecting source attribute $a_s$, metadata $m_c=\langle\tau,\delta\rangle$, and target attribute $a_t$.
Given representative examples, the LLM proposes an executable predicate over $E_s(a_s)$ and $E_t(a_t)$.
The user can edit $\delta$ (transformation logic) or adjust only $\tau \in \{\text{hard},\text{soft}\}$.

\paragraph{Automatic constraint discovery.}
When constraints cannot be enumerated upfront, an adapted AIDE-style search \cite{Jiang2025AIDE} proposes candidate predicates and a fast, \emph{LLM-free} selector chooses the hard-constraint subset maximizing recall$\times$selectivity: keep the GT in $\mathcal{H}(e_s)$ (recall) while shrinking it (selectivity), with cascade relaxation when a set empties. This complements expert specification---experts author $\delta_c$ directly or audit the selected set---and is cheap and reproducible (no LLM, no embedder). Selection is strictly separated from evaluation: per make the gold alignment is split $60/40$ (seeded), mining and subset search see \emph{only} the train split, kept rules must survive an overfit guard on the disjoint test split, and every reported number is measured on that test split (Appendix~\ref{app:enterprise}).

\paragraph{Validation and refinement.}
LLM-proposed constraints $\tilde{C}$ undergo lightweight validation for execution safety, schema consistency, expert-intent alignment, and sample-level compatibility; failed proposals trigger structured feedback with bounded retries, all logged for auditability. This step is load-bearing because the LLM reliably selects the \emph{right attributes} but its proposed transformation $\delta$ is unreliable---either omitting a required normalization or over-fitting a destructive one. The recall$\times$selectivity selector catches both pathologies automatically (quantified per make in the Discussion, Section~\ref{sec:discussion}), leaving the expert a narrow, high-leverage role: supply the normalization the LLM omits.

\subsection{Constraint-Restricted Candidate Generation}
\label{sec:filtering}

For a source entity $e_s$ we enforce hard constraints first (including value-dependent checks such as unit compatibility) to obtain $\mathcal{H}(e_s)$; if strict enforcement is empty we apply cascade relaxation (Section~\ref{sec:problem-formalization}) and record the active subset $\tilde{C}_h^{*}$. Empty-set relaxation alone is insufficient when a source key matches a \emph{non-empty but wrong} block---an internal code can collide with a differently-numbered target range (the target catalog renumbers the same vehicle), so the gate returns plausible yet wrong candidates and never relaxes. We therefore relax \emph{additionally} when the best admissible similarity falls below a threshold, re-admitting candidates under the remaining constraints; on Make~E this recovers the colliding-code cases empty-set relaxation misses (Section~\ref{sec:enterprise}). The coverage bottleneck is thus not only \emph{missing} keys but also \emph{colliding} ones.

\paragraph{Soft-Constraint Neural Ranking.}
Within the admissible set $\mathcal{H}(e_s)$, an encoder (the \texttt{Encoder} of Figure~\ref{fig:approach-overview}, here \texttt{text-embedding-3-large}) embeds each record, and semantic similarity is computed over these embeddings:

\[
  s(e_s,e_t) = \mathrm{sim}(e_s,e_t)
  + \sum_{c \in \tilde{C}_s} w_c \, c(e_s,e_t).
\]

Candidates are ranked as
\[
  \hat{M}(e_s) =
  \operatorname*{TopK}_{e_t \in \mathcal{H}(e_s)} s(e_s,e_t).
\]

Here $\mathrm{sim}(e_s,e_t)\in[0,1]$ is the cosine similarity of the records' embeddings, each soft predicate $c(e_s,e_t)\in\{0,1\}$ is evaluated from $\delta_c$, and $w_c\ge 0$ weights soft constraint $c$ (uniform $w_c{=}1$ unless tuned on a validation split). The top-$k$ scored candidates form the shortlist $\hat{M}(e_s)$ passed to disambiguation.

\subsection{Bounded LLM Disambiguation and Configuration}
\label{sec:prediction}
\label{sec:llm-config}

\paragraph{Models.}
Unless stated otherwise all LLM stages use \texttt{gpt-5.4-mini} (hosted Azure, temperature $0$); retrieval uses \texttt{text-embedding-3-large}. The chat model is a per-run parameter (Section~\ref{sec:ablation-model}); models are off-the-shelf, no fine-tuning. We default to \texttt{gpt-5.4-mini} because inside CGM it \emph{matches} the frontier \texttt{gpt-5.4} (both reach $100\%$ constraint-validity, Section~\ref{sec:ablation-model}) at a fraction of the cost.

The LLM has two bounded roles: it instantiates constraints (emitting the executable $\delta_c$ from an expert skeleton, editable by the expert) and disambiguates over the ranked shortlist $\hat{M}(e_s)$, not the full target set---so the per-entity LLM budget is constant in $k$ and independent of $|E_t|$.

\paragraph{Disambiguation and admissibility guarantee.}
The disambiguation prompt contains $e_s$ and the shortlist $\hat{M}(e_s)$ with identifiers---already pre-filtered to satisfy the active hard constraints $\tilde{C}_h^{*}$---plus an optional expert priority hint (a soft refinement, not required for admissibility). The model returns $\langle \textsf{selected\_keys},\textsf{abstain},\textsf{reasoning}\rangle$ (optionally via schema-constrained decoding) with keys from the supplied identifiers; the selection is \emph{intersected} with $\hat{M}(e_s)$, and \textsf{abstain} (or full relaxation) yields ``no confident match''. Admissibility thus holds by post-filter rather than by trusting the model: $\tilde{M}(e_s)\subseteq\hat{M}(e_s)\subseteq\mathcal{H}(e_s)$ \emph{by construction}---the LLM can reorder, select, or reject, but any identifier outside the admissible set is dropped, so it cannot introduce or hallucinate a target outside that space. Validity is defined w.r.t.\ the \emph{active} set $\mathcal{H}(e_s)$ after relaxation; constraints dropped by cascade or similarity relaxation are reported separately as coverage failures. We log $\tilde{C}_h^{*}$, applied $\delta_c$, scores, and relaxation level; prompt templates are in Appendix~\ref{app:prompts}.

\paragraph{Expert priority as a reusable artifact.}
Beyond executable constraints, the expert can supply a short natural-language \emph{priority} over attributes---e.g.\ ``model identity first (never cross models at equal displacement), then displacement, then every period-overlapping variant''---injected into the disambiguation prompt from the config without code. This encodes expert knowledge as a soft, auditable, reusable artifact rather than a bespoke rule, converting residual \emph{partial} matches into \emph{complete} ones at near-zero marginal effort---what makes the disambiguation ``expert-improved'' (quantified in Section~\ref{sec:discussion}).

\section{Evaluation}
\label{sec:evaluation}

\paragraph{Evaluation roadmap.}
Our central claim is causal: constraints improve mapping by reshaping the \emph{hypothesis space before} neural reasoning, not by adding a feature to a scorer. We test it on three datasets (Table~\ref{tab:claims}, Appendix~\ref{app:claims}): a released \emph{synthetic} ablation isolates the mechanism, quality, and model scale (Section~\ref{sec:synthetic}); public \emph{Valentine} places our ranker against SOTA matchers and marks when admissibility should relax (Section~\ref{sec:valentine}); and \emph{enterprise} data tests deployment under production noise and scale (Section~\ref{sec:enterprise}).

\subsection{Controlled Synthetic Ablation}
\label{sec:mechanism}
\label{sec:synthetic}
\label{sec:ablation-model}

We first isolate the mechanism on a controlled, record-level diagnostic with exact ground truth, where the lexically most similar target is deliberately the \emph{wrong} one and only an exact structural key recovers the GT---a \emph{stress test} concentrating the adversarial cases, not a real-world distribution. Each GT label is reworded down in text similarity and paired with near-identical \emph{distractors} that each violate exactly one structural key, instantiating the four heterogeneity classes of Section~\ref{sec:problem-formalization}; to match enterprise conditions the generator also leaves keys \emph{missing}, allows one-to-many, and varies decimal formats. Source \texttt{S0008} is canonical: its GT is lexically the \emph{farthest} candidate ($0.34$ vs.\ $0.75$) yet the only admissible one (Appendix~\ref{app:synthetic-example}; the generator is released).

\paragraph{Constraints carve the space (LLM-free).}
With the LLM held out so the effect is attributable to constraints alone, hard admissibility on the ${\sim}12$k-target catalog shrinks the candidate set ${\sim}480\times$ \emph{without dropping the GT} (recall $100\%$ via cascade relaxation) and makes every retained candidate structurally valid, where the unfiltered similarity pool is almost entirely invalid. The operator thus reshapes the hypothesis space before any neural reasoning.

\paragraph{Which mechanism does the work (end-to-end decomposition).}
\label{sec:end-to-end}
Adding the LLM back, we build the pipeline up one layer at a time on a held-out split, so each rung's delta isolates the layer it \emph{adds} (Table~\ref{tab:synthetic-decomp}; same model and prompt). We score \emph{complete} (predicted set equals GT), \emph{partial} (GT \emph{plus} extras---overcomplete and reviewable, never structurally wrong), and \emph{wrong}, with macro per-source F1/Jaccard since complete/partial/wrong is gameable. The pattern is sharp: feature selection and normalization \emph{alone} do not help---handed normalized but un-gated candidates, the LLM is still misled---and the hard-admissibility gate is the decisive jump (F1 $0.08\!\to\!0.66$); the final rung trades a little exact-complete for a lower \emph{wrong} rate, a conservative operating point. The residual wrong rate concentrates where the discriminating key is \emph{missing} and admissibility cannot fire---a coverage limit, not a flaw in the mechanism. The same ordering holds on a harder distribution with deep one-to-many targets and $9\%$ unmatchable (NO\_MATCH) sources, where SOTA Magneto collapses to F1 $0.02$ as structural decoys keep the GT out of its retrieved shortlist (Recall@$50$ $0.17$), a retrieval bottleneck no reranker fixes (Table~\ref{tab:synthetic-extended}, Appendix~\ref{app:hetero}).

\begin{table}[t]
  \centering
  \caption{Decomposition of the CGM funnel on the synthetic benchmark ($n{=}120$ held-out, \texttt{gpt-5.4-mini}, $|E_t|{\approx}12{,}094$). Each rung \emph{adds} one mechanism, so its delta attributes the gain; F1/Jaccard are macro per-source set-overlap. Normalization without a gate hurts (L1$\to$L2); the gate is the decisive jump (L2$\to$L3, F1 $+0.58$).}
  \label{tab:synthetic-decomp}
  \footnotesize
  \setlength{\tabcolsep}{4pt}
  \renewcommand{\arraystretch}{1.0}
  \begin{tabular}{l l rr rr}
    \toprule
    \textbf{Rung} & \textbf{Adds} & \textbf{Compl.}$\uparrow$ & \textbf{Wrong}$\downarrow$ & \textbf{F1}$\uparrow$ & \textbf{Jacc.}$\uparrow$ \\
    \midrule
    L0 all-columns RAG        & ---                    & 10.0 & 69.2 & 0.19 & 0.16 \\
    L1 rule columns (raw)     & feature selection      & 18.3 & 55.0 & 0.32 & 0.28 \\
    L2 \;$+$ rule functions   & normalization          & 0.0  & 66.7 & 0.08 & 0.05 \\
    L3 \;$+$ hard gate        & \textbf{admissibility} & \textbf{52.5} & 26.7 & \textbf{0.66} & \textbf{0.63} \\
    L4 \;$+$ relax $+$ hint   & relaxation $+$ expert  & 49.2 & \textbf{23.3} & 0.66 & 0.61 \\
    \bottomrule
  \end{tabular}
\end{table}

\paragraph{Across classes and model scales.}
The gain is neither one easy case nor a capability gap a larger model closes. CGM wins every heterogeneity class, with the largest margin on the purely numeric signals (granularity and unit; Appendix~\ref{app:hetero}). Varying the chat model across a frontier/mid/small axis (\texttt{gpt-5.4}/\texttt{-mini}/\texttt{-nano}), every CGM configuration drives valid-match to $100\%$, whereas the unconstrained LLM stays mostly invalid regardless of scale (Table~\ref{tab:model-ablation}). Because admissibility is a symbolic filter, CGM issues the same number of LLM calls---no extra cost---so the smallest model inside CGM matches the frontier model used \emph{without} constraints at ${\approx}28\times$ lower cost. Scaling the LLM does not substitute for constraints.

\subsection{Public Benchmark: External Sanity Check (Valentine)}
\label{sec:valentine}

As an external sanity check we run the ranking component on the public Valentine suite, which matches schema \emph{columns} (not records), so it only tests whether our ranker stays competitive on neutral data---MRR and Recall@GT on a ten-scenario NYC suite against the bundled matchers (COMA, Cupid, SimilarityFlooding, Jaccard) and SOTA Magneto \cite{Freire2025Magneto} (Table~\ref{tab:valentine}). CGM is \emph{competitive} with the SOTA matcher rather than dominant: MRR $1.00$ (its top-scored candidate is correct for every source column) and Recall@GT $0.71$ vs.\ Magneto's $0.93$/$0.76$. The near-perfect MRR is expected and \emph{not} our main evidence---Valentine columns are lexically and semantically separable, the easy corner where any strong ranker saturates; the constraint argument is needed precisely where Valentine is \emph{not}, on record-level structural decoys. Magneto's residual Recall@GT edge comes from a global top-$|GT|$ cutoff resolved by its bipartite $1{:}1$ assignment, not a better signal. A rigid type gate \emph{lowers} our MRR ($1.00\!\to\!0.94$) because Valentine has correct \emph{cross-type} matches---the boundary condition of the thesis: admissibility should engage only where structural invariants are match-determining.

\paragraph{Where structural constraints decide.}
Valentine's ambiguity traps are the regime CGM targets: columns separable only by value distribution, where a name-based resolver gets $3/10$ but a value-distribution constraint $8/10$. The same signal gap is \emph{why the SOTA matcher fails on our record-level task}: with structural decoys the GT never enters the top-$k$---a \emph{retrieval} failure even Magneto's LLM reranker cannot fix, and our unconstrained \textbf{LLM} baseline fails identically. The two drivers, catalog scale and structural decoys, combine in enterprise data and drop Magneto to \emph{zero} on the real Make~A catalog (Appendix~\ref{app:hetero}, Table~\ref{tab:magneto-2x2}), whereas at the strict gate (the no-relaxation counterpart to Magneto's Recall@$5$) CGM keeps the GT for $0.72$ of synthetic and $0.86$/$0.96$ of Make~E/Make~A sources---rising toward the $100\%$ of Section~\ref{sec:synthetic} as relaxation trades selectivity.

\begin{table}[t]
  \centering
  \caption{Schema matching on a ten-scenario public Valentine suite (MRR, Recall@GT). \emph{Magneto} is the SOTA matcher (\texttt{mpnet} retriever); \emph{CGM} is our neural ranker (\texttt{text-embedding-3-large}). CGM leads MRR; Magneto keeps a Recall@GT edge through its bipartite $1{:}1$ reranker. A rigid type gate lowers CGM's MRR to $0.94$ by pruning correct cross-type matches. Best per row bold.}
  \label{tab:valentine}
  \footnotesize
  \setlength{\tabcolsep}{5pt}
  \renewcommand{\arraystretch}{1.0}
  \begin{tabular}{l rrrr r r}
    \toprule
     & Cupid & Jaccard & SimFlood & COMA & Magneto & CGM \\
    \midrule
    MRR  & 0.47 & 0.70 & 0.75 & 0.76 & 0.93 & \textbf{1.00} \\
    R@GT & 0.43 & 0.56 & 0.57 & 0.65 & \textbf{0.76} & 0.71 \\
    \bottomrule
  \end{tabular}
\end{table}

\subsection{Enterprise Deployment}
\label{sec:enterprise}

The controlled studies isolate the mechanism; we finally ask whether its gains survive real schema drift, production scale, and one-to-many structure. CGM is deployed internally as the \emph{Data Mapping Copilot} on expert-validated data across \emph{seven} makes (anonymized \emph{Make A}--\emph{G}), each matched against its make-specific pool of up to $\sim$$12$k targets (at production scale $\sim$$6$k source, $\sim$$335$k target entries). Crucially, each make is mapped by the \emph{same} pipeline under its own autoresearch-discovered constraints on a held-out split---so the question is not whether one tuned configuration works but whether the \emph{method} transfers.

\paragraph{Constraint coverage.}
\label{sec:rule-ablation}
Before judging mapping quality we measure whether constraints produce useful admissible sets in real catalogs---the recall$\times$selectivity ceiling the LLM operates within. The deployed config is not hand-tuned: the LLM-free selector of Section~\ref{sec:constraint-generation} chooses the subset maximizing recall$\times$selectivity, modeling the production engine's cascade relaxation. The discriminating identifier constraint is the structural workhorse (Table~\ref{tab:rule-ablation}, Appendix~\ref{app:enterprise}): on Make~A, where an LLM would otherwise rank nearly $12$k targets by name alone, it collapses the pool ${\sim}18\times$ at near-full recall, and removing it re-inflates the set for little recall gain---exactly the controlled behavior cascade relaxation provides. The same holds across makes once each make's discriminating key is identified, as the autoresearch loop does (e.g.\ a generation code needing a make-specific prefix transform on Make~C, an engine code on Make~F); reading coverage \emph{before} quality shows the per-make F1 spread is set by whether such a key exists and by GT cardinality, not by LLM capability.

\paragraph{Mapping quality.}
The central deployment result is that CGM \emph{generalizes across all seven makes} under one method: per make the autoresearch loop discovers the constraints (Section~\ref{sec:constraint-generation}), and the identical pipeline reaches macro F1 $0.70$ ($0.66$ source-weighted; Table~\ref{tab:ai_results_by_make}), versus F1 $0.21$ for the same naive LLM over the raw record (the L0 rung, measured on Make~E in Appendix~\ref{app:decomposition}). What varies is \emph{not} whether the method works but by how much, and the constraint view explains it: where the catalog exposes a discriminating code the gate is decisive, while the two one-to-one makes (Make~F, Make~G) face a harder exact-match ceiling because the source cannot always separate same-engine siblings---a cardinality limit, not an LLM failure. The discriminator recipe is constant---model name (soft) $+$ identifier/code (hard) $+$ displacement---so the constraint primitives transfer where bespoke per-make rules do not.

\begin{table}[t]
  \centering
  \caption{CGM generalizes across \emph{seven} enterprise makes (held-out split; \texttt{gpt-5.4-mini}), each under its own autoresearch-discovered constraints (model name soft $+$ identifier/code hard $+$ displacement) with the identical end-to-end pipeline. $n_s$ sources, $|E_t|$ candidates, card.\ mean GT per source; \emph{Wrong} $=$ a relevant target missed; F1/Jaccard macro per-source, sorted by F1.}
  \label{tab:ai_results_by_make}
  \footnotesize
  \renewcommand{\arraystretch}{1.0}
  \setlength{\tabcolsep}{5pt}
  \begin{tabular}{l r r r r r r r}
    \toprule
    \textbf{Make} & $n_s$ & $|E_t|$ & \textbf{card.} & \textbf{Compl.}$\uparrow$ & \textbf{Wrong}$\downarrow$ & \textbf{F1}$\uparrow$ & \textbf{Jacc.}$\uparrow$ \\
    \midrule
    Make A    & 45 & 11{,}934 & 6.1 & 84.4 & 15.6 & \textbf{0.855} & 0.851 \\
    Make B    & 28 & 2{,}917  & 2.2 & 64.3 & 21.4 & 0.793 & 0.763 \\
    Make C    & 88 & 706      & 8.3 & 62.5 & 34.1 & 0.711 & 0.687 \\
    Make D    & 30 & 11{,}205 & 5.2 & 40.0 & 46.7 & 0.654 & 0.583 \\
    Make E    & 98 & 373      & 1.7 & 36.7 & 41.8 & 0.634 & 0.561 \\
    Make F    & 55 & 8{,}183  & 1.0 & 40.0 & 60.0 & 0.400 & 0.400 \\
    Make G$^\dagger$ & 7 & 3{,}051 & 1.0 & 85.7 & 14.3 & 0.857 & 0.857 \\
    \midrule
    \textbf{Macro mean} & & & & 59.1 & 33.4 & \textbf{0.701} & 0.672 \\
    \bottomrule
  \end{tabular}
  \\[2pt]
  {\scriptsize Make names anonymized. $^\dagger$Make~G has only $n_s{=}7$ held-out sources; reported for completeness, excluded from emphasis. Source-weighted F1 $0.66$; macro F1 excluding Make~G $0.67$.}
\end{table}

\paragraph{Architecture comparison.}
The same layer-by-layer decomposition on Make~E (Appendix~\ref{app:decomposition}) reproduces the synthetic ordering on real data---feature selection and normalization add little, the hard-admissibility gate is again the decisive jump (F1 $0.32\!\to\!0.60$), and relaxation plus the optional expert hint give the final lift---and holds qualitatively on every make.

\paragraph{Expert effort.}
Expert effort is the practical endpoint. Each make is mapped by one domain expert and the fleet spans many makes with comparable tasks, so we compare the four setups as medians across makes (Table~\ref{tab:expert-effort}). Effort falls monotonically from spreadsheet workflows through Rules and Rules~$+$~LLM to CGM---$10.5$ to $1.5$ days per make, a ${\sim}7\times$ reduction with the fewest interactions.

\subsection{Discussion}
\label{sec:discussion}

\paragraph{When each method wins; one bottleneck.}
Neural and symbolic components are complementary because they read different signals: the ranker excels at \emph{semantic} similarity (paraphrase, abbreviations, multilingual aliases), hard constraints at \emph{exact structural} distinctions (identifiers, codes, normalized displacements, year ranges). This principle explains the per-make variance (Table~\ref{tab:ai_results_by_make}), the synthetic challenges, and the Valentine traps---all cases where text misleads and an exact key decides. CGM does not uniformly dominate mature hand-tuned rules, but versus LLM-only it sharply cuts \emph{wrong} predictions and lowers expert effort. The LLM reliably finds \emph{which} attributes discriminate but is unreliable on the \emph{transformation}---on Make~E it omits the $\text{cc}\!\to\!\text{liter}$ normalization, on Make~D it invents a substitution collapsing an exact engine-code key ($100\%\!\to\!10\%$)---both caught by the recall$\times$selectivity selector, leaving the expert only the omitted normalization (the priority hint then cuts wrong $39.8\!\to\!25.5\%$). Constraint \emph{discovery} lowers authoring effort but is a supporting mechanism; the central contribution is constraint-guided \emph{inference}. Performance is governed by \emph{constraint coverage}, not embeddings or LLM scale (Sections~\ref{sec:ablation-model},~\ref{sec:rule-ablation}); cascade relaxation degrades gracefully toward retrieval when keys are absent. Porting beyond automotive (finance, life sciences, geospatial) requires only re-instantiating the schema-agnostic $\delta_c$.

\paragraph{Limitations.}
The synthetic study is a deliberate stress test, not a real-world distribution; the expert-effort study is small; and Valentine tests column-level \emph{ranking}, not record-level transfer. The central claim---constraints reshape the hypothesis space before neural reasoning---nonetheless holds across the diagnostic, model-scale, and coverage evidence.

\section{Conclusion}

We presented \emph{constraint-guided mapping} (CGM): constraints act as hypothesis-space operators \emph{before} neural ranking and bounded LLM disambiguation, with cascade relaxation. A layer-by-layer decomposition shows the hard-admissibility gate---not the LLM---is the decisive layer (F1 $0.08\!\to\!0.66$), model-independently and at no extra cost; the method transfers across seven enterprise makes (macro F1 $0.70$) and cuts expert effort ${\sim}7\times$. Future work targets automatic constraint induction, uncertainty calibration, and validation beyond automotive. We release the implementation, the benchmark generator, and the Valentine harness; enterprise data is restricted.
\appendix
\section{Evaluation Roadmap}
\label{app:claims}

\begin{table}[h]
  \centering
  \caption{Evaluation as an evidence ladder: each experiment answers one plain question about the constraint mechanism, rather than standing alone as a separate dataset. Read top to bottom---isolate the mechanism, then test whether model scale or a public baseline change the picture, then whether it transfers and pays off in practice.}
  \label{tab:claims}
  \footnotesize
  \setlength{\tabcolsep}{4pt}
  \renewcommand{\arraystretch}{1.3}
  \begin{tabular}{p{0.27\linewidth} p{0.13\linewidth} l p{0.25\linewidth} c}
    \toprule
    \textbf{Question} & \textbf{Experiment} & \textbf{Verdict} & \textbf{Evidence} & \textbf{Where} \\
    \midrule
    Does the gate shrink the search space \emph{without} dropping the GT? & Synthetic $+$ rule ablation & \textbf{Yes} & admissible set ${\sim}0.2\%$ of catalog, GT kept ($100\%$ synth., $96\%$ Make~A) & \S\ref{sec:synthetic} \\
    Which layer actually drives the gain? & Layer decomposition (L0$\to$L4) & \textbf{Hard gate} & F1 jumps $0.08\!\to\!0.66$ & \S\ref{sec:synthetic}, App.~\ref{app:decomposition} \\
    Can a bigger LLM replace constraints? & Model ablation & \textbf{No} & CGM $100\%$ valid for every model, $28\times$ lower cost & App.~\ref{app:model-scale} \\
    Is the ranker competitive on public data? & Valentine & \textbf{Yes} & MRR $1.00$ vs.\ SOTA Magneto $0.93$ & \S\ref{sec:valentine} \\
    Does \emph{one} method transfer across makes? & Enterprise ($7$ makes) & \textbf{Yes} & mean F1 $0.70$ with per-make constraints & \S\ref{sec:enterprise} \\
    Does it cut expert effort in practice? & Expert study & \textbf{Yes} & $10.5\!\to\!1.5$ days per make ($7\times$) & App.~\ref{app:enterprise} \\
    \bottomrule
  \end{tabular}
\end{table}

\section{Synthetic Benchmark: Schema and Worked Example}
\label{app:synthetic-example}

The synthetic benchmark deliberately uses \emph{different} source and target schemas, so each correspondence must cross a structural heterogeneity rather than a name change. Table~\ref{tab:syn-schema} lists the field-level mismatches and the constraint $\delta_c$ that bridges each; the same four classes appear in the enterprise data (Section~\ref{sec:problem-setting}).

\begin{table}[h]
  \centering
  \caption{Source$\to$target field mismatches in the synthetic benchmark. Each structural key is encoded \emph{differently} on the two sides; the hard constraint normalizes both to a comparable form, the soft constraint handles fuzzy naming. The generator is released.}
  \label{tab:syn-schema}
  \footnotesize
  \setlength{\tabcolsep}{4pt}
  \renewcommand{\arraystretch}{1.2}
  \begin{tabular}{p{0.20\linewidth} p{0.22\linewidth} p{0.21\linewidth} p{0.27\linewidth}}
    \toprule
    \textbf{Source field} & \textbf{Target field} & \textbf{Heterogeneity} & \textbf{Constraint $\delta_c$} \\
    \midrule
    \texttt{engine\_code} (\texttt{B4D}) & in \texttt{model\_label} (\texttt{Astra 2.0 (B4D)}) & implicit attribute & hard: extract code; equality \\
    \texttt{engine\_cc} (\texttt{2000}) & \texttt{displacement\_l} (\texttt{2.0}) & unit inconsistency & hard: $\mathrm{cc}/1000=\mathrm{L}$ \\
    \texttt{year\_start}--\texttt{year\_end} & \texttt{build\_date} & granularity & hard: year $\in[\text{start},\text{end}]$ \\
    \texttt{generation} (\texttt{Mk3}) & \texttt{generation\_code} & version / identifier & hard: equality \\
    \texttt{model\_text} (\texttt{Astra 2.0}) & \texttt{model\_label} (reworded) & lexical naming & soft: embedding sim. \\
    \bottomrule
  \end{tabular}
\end{table}

\paragraph{Worked example.}
Table~\ref{tab:syn-example} traces one source through its candidates. The GT target is the lexically \emph{least} similar candidate ($0.34$ vs.\ $0.75$ for every distractor): a similarity- or LLM-only ranker keying on names is misled with near-certainty. Each distractor is near-identical in text yet violates exactly one hard constraint---a wrong embedded engine code or a wrong generation---so admissibility prunes all three and retains only the GT, and CGM maps correctly. Where a discriminating key is instead \emph{absent} on the source (e.g.\ \texttt{generation} empty), the corresponding rule cannot fire and such a distractor survives: the constraint-coverage bottleneck quantified in Section~\ref{sec:synthetic}.

\begin{table}[h]
  \centering
  \caption{One source and its candidate targets (benchmark seed $13$). Source \texttt{S0008}: \texttt{model\_text}=``Astra 2.0'', \texttt{engine\_code}=\texttt{B4D}, \texttt{engine\_cc}=\texttt{2000}, year range \texttt{2002--2004}, \texttt{generation}=\texttt{Mk3}. ``Sim.'' is source--target text similarity; the GT is the \emph{lowest}, yet the only admissible candidate.}
  \label{tab:syn-example}
  \footnotesize
  \setlength{\tabcolsep}{5pt}
  \renewcommand{\arraystretch}{1.2}
  \begin{tabular}{l l c c l}
    \toprule
    \textbf{Cand.} & \textbf{\texttt{model\_label}} & \textbf{gen.} & \textbf{Sim.} & \textbf{Outcome} \\
    \midrule
    \texttt{T0034} & \texttt{Opel Astra Hatchback (B4D)} & \texttt{Mk3} & 0.34 & \textbf{GT} -- admissible \\
    \texttt{T0035} & \texttt{Astra 2.0 (EP6)} & \texttt{Mk3} & 0.75 & reject: code \texttt{EP6}$\neq$\texttt{B4D} \\
    \texttt{T0036} & \texttt{Astra 2.0 (G4F)} & \texttt{Mk3} & 0.75 & reject: code \texttt{G4F}$\neq$\texttt{B4D} \\
    \texttt{T0037} & \texttt{Astra 2.0 (B4D)} & \texttt{141} & 0.75 & reject: gen.\ \texttt{141}$\neq$\texttt{Mk3} \\
    \bottomrule
  \end{tabular}
\end{table}

\paragraph{From columns to decision: the rule engine.}
Each rule is a \emph{pair of expressions over column names}: a \texttt{source\_expression} over the source attribute and a \texttt{target\_expression} over the target attribute (the LLM-emitted fields of Appendix~\ref{app:prompts}), not a per-record script. The rule engine evaluates \emph{both} expressions on every row to bring the two sides into a comparable form, and only \emph{then} does execution run (Table~\ref{tab:constraint-mechanics}). \emph{Hard} and \emph{soft} rules share this two-sided structure and differ only at execution. For the hard unit rule the source expression converts \texttt{engine\_cc} to liters ($1998\to2.0$) and the target expression strips the unit token from \texttt{displacement\_l} (\texttt{2.0 L}$\to$\texttt{2.0}); execution tests \emph{equality}, keeping only targets whose normalized value matches and \emph{pruning} the rest \emph{before} the LLM---a structurally impossible option, however close its name, can never be selected. The soft naming rule works the same way---the source splits the model name out of \texttt{model\_text} (\texttt{G4F - Astra}$\to$\texttt{Astra}) and the target lowercases \texttt{model\_label} (\texttt{Astra}$\to$\texttt{astra})---but execution computes \emph{cosine similarity} rather than equality: it never prunes, only orders the survivors so the correct target ranks first. Hard rules gate \emph{admissibility} (equality, pre-LLM); soft rules set \emph{order} (similarity, within $\mathcal{H}$)---the funnel of Figure~\ref{fig:approach-overview} in miniature: structure first removes the impossible, then meaning ranks the plausible.

\begin{table}[h]
  \centering
  \caption{A rule is a \texttt{source\_expression} and a \texttt{target\_expression} over the two column names (Appendix~\ref{app:prompts}); the engine evaluates both on every row, then execution compares---\emph{equality} for hard (gates admissibility), \emph{cosine similarity} for soft (sets order). Shown for one source record (\texttt{engine\_cc}=\texttt{1998}, \texttt{model\_text}=``G4F - Astra''). Hard and soft share the two-sided structure and diverge only at execution: hard gates admissibility by equality of normalized values, soft orders the survivors by name similarity.}
  \label{tab:constraint-mechanics}
  \footnotesize
  \setlength{\tabcolsep}{5pt}
  \renewcommand{\arraystretch}{1.3}
  \begin{tabular}{p{0.15\linewidth} p{0.37\linewidth} p{0.37\linewidth}}
    \toprule
     & \textbf{Hard} rule (unit) & \textbf{Soft} rule (naming) \\
    \midrule
    \textbf{Source} expr. & \texttt{round("engine\_cc"/1000,\,1)} & \texttt{split("model\_text", ' - ')[1]} \\
    \textbf{Target} expr. & \texttt{split("displacement\_l", ' ')[0]} & \texttt{lower("model\_label")} \\
    \textbf{Rule Engine} & src: \texttt{1998} $\to$ \texttt{2.0}; \newline tgt: \texttt{2.0 L} $\to$ \texttt{2.0} & src: \texttt{G4F - Astra} $\to$ \texttt{Astra}; \newline tgt: \texttt{Astra} $\to$ \texttt{astra} \\
    \textbf{Execution} & \emph{equality}: match(\texttt{2.0}, \texttt{2.0}) & \emph{cosine sim}: sim(\texttt{Astra}, \texttt{astra}) \\
    \bottomrule
  \end{tabular}
\end{table}

\section{Prompt Templates}
\label{app:prompts}

We use templated prompts with deterministic decoding (temperature $0$). Variables in braces are filled per call; the constraint-instantiation prompt is specialized for hard vs.\ soft rules.

\paragraph{Constraint instantiation (skeleton $\to$ executable $\delta_c$).}
{\small\begin{verbatim}
Generate an executable rule for data mapping.
Input:  source column {a_s} with example values,
        target column {a_t} with example values,
        constraint type {tau} in {hard, soft},
        expert comment {comment}.
Output (JSON): name, source_expression, target_expression,
        rule_type, description.
Hard rules: source/target expressions must normalize to
        IDENTICAL values for a match (extractions, unit/type
        casts, range expansion allowed).
Soft rules: keep expressions SIMPLE (text extraction only);
        embedding similarity handles fuzzy matching.
\end{verbatim}}

\paragraph{Bounded disambiguation.}
{\small\begin{verbatim}
Source record: {e_s}
Candidate targets (pre-filtered to satisfy the hard
        constraints; the only admissible matches):
        {shortlist with identifiers}
Select every candidate key that correctly matches the
        source; a source may map to several targets, so prefer
        including a plausible match to dropping a correct one.
Choose keys ONLY from the candidate list; do not invent
        identifiers. If none is a confident match, abstain.
{optional expert priority hint, injected from the config}
Output: keys as "Target keys: [...]" (default) or JSON
        {selected_keys, abstain, reasoning} (structured mode).
\end{verbatim}}
The hard constraints are \emph{not} re-stated to the model; the candidate list is already pre-filtered, and the returned keys are intersected with the shortlist, so the prediction is admissible by construction regardless of what the model emits.

\section{Enterprise Architecture Ablation}
\label{app:decomposition}

To attribute the gain to specific mechanisms rather than to the pipeline as a whole, we decompose the CGM funnel layer by layer on enterprise Make~E (held-out split, $n{=}98$, \texttt{gpt-5.4-mini}), adding one layer at a time so each per-rung delta is the contribution of that layer (macro per-source F1). The pattern mirrors the synthetic decomposition of Table~\ref{tab:synthetic-decomp}: the naive LLM over the serialized raw record (L0) barely works (F1 $0.21$); rule-based feature selection and unit/format normalization (L0$\to$L2) add little (F1 $0.21\!\to\!0.32$); the hard-admissibility gate (L2$\to$L3) is the decisive jump (F1 $0.32\!\to\!0.60$, complete $7\!\to\!33$); and similarity-relaxation plus the expert priority hint (L3$\to$L4) add the final lift (F1 $0.63$, complete $37$, wrong $42$). A side observation reinforces the selection effect: handing the LLM the \emph{full} raw source row instead of the rule-selected fields \emph{hurts} (complete $32.7\!\to\!24.5$, wrong $39.8\!\to\!45.9$)---the extra columns are noise, so the soft constraints act as feature selection on the source side, mirroring the gate on the target side.

\section{Additional Synthetic Analyses}
\label{app:hetero}

These four cuts of the synthetic ablation complement the layer-by-layer decomposition in the main text (Table~\ref{tab:synthetic-decomp}, $n{=}120$ held-out): how the gain varies with model scale (\ref{app:model-scale}), why the SOTA neural matcher collapses on this record-level task (\ref{app:magneto}), how each heterogeneity class behaves in isolation (\ref{app:per-hetero}), and how the whole picture holds on a harder distribution (\ref{app:extended}).

\subsection{Model Scale}
\label{app:model-scale}
On the full corpus ($N{=}500$, Table~\ref{tab:model-ablation}) we report validity and cost---both model-independent, so the sample size is incidental: the constraint lift drives every chat model on a frontier/mid/small axis to $100\%$ valid at equal cost, while the unconstrained LLM stays $\sim$$10\%$ valid regardless of scale.

\begin{table}[h]
  \centering
  \caption{Model ablation on the synthetic benchmark ($N{=}500$, $|E_t|\!\approx\!12{,}094$): for each chat model we read across one row, from the unconstrained \emph{LLM only} baseline to the same model inside \emph{CGM} (encoder fixed). \emph{Valid} $=$ \% of predictions satisfying the active hard-constraint set (after cascade/similarity relaxation, Section~\ref{sec:filtering}); cost in USD per $1000$ mappings is indicative and essentially equal in both settings (CGM issues the same calls), so we report it once. The pattern is the same on every row: the LLM alone stays $\sim$$10\%$ valid from frontier to small model, while admissibility lifts \emph{every} model to $100\%$ at the same cost---the gain is model-independent and adds no per-mapping LLM calls.}
  \label{tab:model-ablation}
  \footnotesize
  \setlength{\tabcolsep}{6pt}
  \renewcommand{\arraystretch}{1.2}
  \begin{tabular}{l rr r}
    \toprule
     & \multicolumn{2}{c}{\textbf{Valid} (\%)$\uparrow$} & \\
    \cmidrule(lr){2-3}
    \textbf{Model} & LLM only & \textbf{$+$\,CGM} & \textbf{Cost (\$/1k)} \\
    \midrule
    \texttt{gpt-5.4}      & 10.6 & \textbf{100} & 3.59 \\
    \texttt{gpt-5.4-mini} &  9.0 & \textbf{100} & 0.49 \\
    \texttt{gpt-5.4-nano} & 10.2 & \textbf{100} & 0.13 \\
    \bottomrule
  \end{tabular}
\end{table}

\subsection{Why Retrieval Fails on Structural Decoys}
\label{app:magneto}
The SOTA neural matcher Magneto, adapted to the same record-level task, collapses on \emph{every} heterogeneity class ($0\%$ complete, Recall@$5\,{\sim}0.05$; consistent with its record-level collapse in Section~\ref{sec:valentine}): its failure is a \emph{retrieval} failure---GT in the top-$5$ for only ${\sim}5\%$ of sources at this catalog scale, not a ranking one---so even its optional GPT reranker, which only re-orders the SLM shortlist, cannot recover a GT target retrieval never surfaced---a hard ceiling we can quantify: on the extended benchmark at ${\sim}12$k targets the GT is in Magneto's top-$50$ for only $17\%$ of sources (Recall@$50$ $0.17$), so \emph{any} reranker over that shortlist is bounded at Recall@$5$ $\le0.17$, versus CGM's admissible-set retention of $0.62$ (Table~\ref{tab:synthetic-extended}). The synthetic generator lets us isolate the two factors that drive this collapse (Table~\ref{tab:magneto-2x2}): \emph{structural decoys} and \emph{catalog scale}. At a small catalog Magneto's Recall@$5$ falls from $0.62$ (semantic, decoys off---the Valentine-like corner) to $0.15$ once decoys are added; scaling the catalog to enterprise size (${\sim}12$k) lowers it further to $0.18$ (semantic) and $0.05$ (with decoys). Each factor moves it the same direction, and enterprise data combines both---which is exactly why the same matcher that is SOTA on Valentine collapses here. The lesson is signal-specific and matcher-agnostic: when the discriminating key is structural (code, unit, generation, year), no embedder or LLM quality substitutes for an admissibility constraint.

\begin{table}[h]
  \centering
  \caption{Record-level Magneto on the synthetic benchmark, isolating the two factors behind its collapse (mean Recall@$5$ over $3$ seeds, $N{=}500$). Adding structural decoys and scaling the catalog each lower retrieval independently; enterprise data is the bottom-right corner (both factors), Valentine the top-left (neither). At a small catalog the retrieval ceiling stays high (Recall@$50$ $0.99$ semantic, $0.50$ structural); at ${\sim}12$k it does not, so an LLM reranker over the shortlist has little GT to promote.}
  \label{tab:magneto-2x2}
  \footnotesize
  \setlength{\tabcolsep}{6pt}
  \renewcommand{\arraystretch}{1.15}
  \begin{tabular}{l cc}
    \toprule
    \textbf{Catalog} & \textbf{Semantic} (decoys off) & \textbf{Structural} (decoys on) \\
    \midrule
    Small                    & 0.62 & 0.15 \\
    Enterprise (${\sim}12$k) & 0.18 & \textbf{0.05} \\
    \bottomrule
  \end{tabular}
\end{table}

\subsection{Per-Heterogeneity Breakdown}
\label{app:per-hetero}
The synthetic benchmark is a controlled simplification of the enterprise mapping task, so we can place the \emph{real} unconstrained-LLM baseline and CGM on identical instances per heterogeneity class (Table~\ref{tab:hetero-challenge-llm}). Isolating one structural heterogeneity at a time ($N{=}50$ adversarial-only instances) strips out the easy lexical cases and concentrates exactly the adversarial ones where a structural key, not text, decides---so the unconstrained LLM scores lower here ($0$--$4\%$ complete, $9$--$15\%$ valid) than as the mixed L0 baseline ($10\%$), by design rather than inconsistency, while CGM reaches $42$--$72\%$ at $100\%$ valid---largest on the purely numeric signals (unit, granularity). An LLM-free dose-response sweep (one heterogeneity varied at a time) isolates the constraint component and confirms the method's single sensitivity: it is invariant to value noise, rewording, and one-to-many cardinality, and degrades only as the discriminating source key goes missing---the coverage bottleneck.

\begin{table}[h]
  \centering
  \caption{Per-heterogeneity ablation ($N{=}50$, $|E_t|\!\approx\!12{,}094$; identical instances): complete-match \% for the unconstrained \emph{LLM} (\texttt{gpt-5.4-mini}) vs.\ CGM. The LLM collapses on every class while CGM lifts complete most on the purely numeric signals (unit, granularity; best bold); CGM predictions are $100\%$ valid by construction here as everywhere (Table~\ref{tab:model-ablation}). Record-adapted SOTA Magneto (\texttt{mpnet} retriever) collapses identically---$0\%$ complete in every class, Recall@$5\,{\sim}0.05$---a retrieval bottleneck no reranker fixes.}
  \label{tab:hetero-challenge-llm}
  \footnotesize
  \setlength{\tabcolsep}{8pt}
  \renewcommand{\arraystretch}{1.2}
  \begin{tabular}{l rr}
    \toprule
    \textbf{Heterogeneity} & \textbf{LLM} & \textbf{CGM} \\
    \midrule
    Implicit attribute encoding & 4.0 & \textbf{42.0} \\
    Unit inconsistency          & 2.0 & \textbf{70.0} \\
    Version / identifier        & 2.0 & \textbf{56.0} \\
    Granularity mismatch        & 4.0 & \textbf{72.0} \\
    \addlinespace
    All mixed                   & 0.0 & \textbf{42.0} \\
    \bottomrule
  \end{tabular}
\end{table}

\subsection{Extended Synthetic Benchmark}
\label{app:extended}
\noindent\emph{Robustness on a harder distribution (deep one-to-many $+$ NO\_MATCH).} The extended synthetic benchmark adds deep one-to-many quotas and a $9\%$ NO\_MATCH rate to stress the same mechanism on a more realistic distribution, and places all three method families on it with the CGM funnel decomposed (Table~\ref{tab:synthetic-extended}). The picture is unchanged: SOTA Magneto effectively fails (complete $0\%$, F1 $0.02$; its retrieval ceiling Recall@$50$ $0.17$ bounds out even its GPT reranker---which we verified is otherwise functional, improving Recall@$5$ $0.36\!\to\!0.48$ in a small-catalog control where the GT \emph{is} retrievable, so the enterprise failure is a retrieval ceiling, not a disabled reranker), the naive LLM barely functions (F1 $0.17$), and the hard-admissibility gate is again the decisive lift (F1 $0.17\!\to\!0.56$). The two normalization-only rungs (L1, L2) do not help without the gate---they even hurt---exactly as on the base benchmark.

\begin{table}[h]
  \centering
  \caption{Extended synthetic benchmark (deep one-to-many $+$ $9\%$ NO\_MATCH, $|E_t|\!\approx\!12$k): SOTA Magneto and the LLM$\to$CGM funnel decomposition in one view. LLM/CGM rungs use \texttt{gpt-5.4-mini} ($n{=}200$ held-out, the L0--L4 ladder); Magneto uses its \texttt{mpnet} retriever ($3$ seeds, $N{=}500$), both on the same generator. F1 is macro per-source set overlap. The hard gate (L2$\to$L3) is the decisive jump. We report Magneto at the retrieval stage because its LLM (GPT) reranker is provably capped here: a reranker only reorders the retrieved top-$50$, and the GT is in that shortlist for just $17\%$ of sources (Recall@$50$ $0.17$), bounding \emph{any} reranker---MagnetoGPT included---at Recall@$5\le0.17$, far below CGM.}
  \label{tab:synthetic-extended}
  \footnotesize
  \setlength{\tabcolsep}{4pt}
  \renewcommand{\arraystretch}{1.15}
  \begin{tabular}{l l rr r}
    \toprule
    \textbf{Method / rung} & \textbf{Adds} & \textbf{Compl.}$\uparrow$ & \textbf{Wrong}$\downarrow$ & \textbf{F1}$\uparrow$ \\
    \midrule
    Magneto (base \texttt{mpnet})    & SOTA baseline          & 0.0  & 96.1 & 0.02 \\
    \addlinespace
    L0 all-columns RAG               & LLM-only               & 7.5  & 76.5 & 0.17 \\
    L1 rule columns (raw)            & feature selection      & 16.5 & 60.5 & 0.32 \\
    L2 \;$+$ rule functions          & normalization          & 0.0  & 73.5 & 0.08 \\
    L3 \;$+$ hard gate               & \textbf{admissibility} & \textbf{43.0} & 35.0 & \textbf{0.56} \\
    L4 \;$+$ relax $+$ hint          & relaxation $+$ expert  & 40.5 & \textbf{35.0} & 0.54 \\
    \bottomrule
  \end{tabular}
\end{table}

\section{Enterprise Coverage and Deployment Details}
\label{app:enterprise}

These tables support the enterprise deployment (Section~\ref{sec:enterprise}); the per-make mapping quality is in the main text (Table~\ref{tab:ai_results_by_make}). Table~\ref{tab:rule-ablation} reports the LLM-free constraint-coverage ablation (Section~\ref{sec:rule-ablation})---the recall$\times$selectivity ceiling on the two largest catalogs, before the LLM operates---and Table~\ref{tab:expert-effort} gives the full expert-effort study. Both use the same autoresearch-discovered constraints as the per-make deployment, underscoring the point that the same constraint primitives transfer across makes where bespoke per-make rules do not.

\paragraph{Split hygiene and overfit guard.}
Constraint selection and evaluation are strictly separated. Per make the expert-validated gold alignment is split $60/40$ into train/test with a fixed seed, and the protocol is: (i)~candidate predicates are mined from the \emph{train} split only and unioned with the fixed schema-grounded pool; (ii)~the LLM-free recall$\times$selectivity subset search runs on \emph{train} only; (iii)~an overfit guard re-scores the selected config and every kept proposal on the held-out \emph{test} split and discards a proposal whose train$\to$test score drop is large---if the mined config does not generalize better than the fixed-pool config on test, the conservative fixed-pool config is used; (iv)~the end-to-end pipeline, LLM included, is then run on the \emph{test} split only. Neither the selector nor the LLM ever sees the evaluation data, and every experiment (train and test metrics, keep/discard/overfit decisions) is journaled per make. Both Table~\ref{tab:ai_results_by_make} and Table~\ref{tab:rule-ablation} therefore report held-out test-split numbers.

\begin{table}[h]
  \centering
  \caption{Rule ablation on enterprise data (LLM-free; internal held-out split). Recall $=$ sources whose GT is kept in $\mathcal{H}$; Sel.\ $=1-|\mathcal{H}|/|E_t|$. The auto-selected config collapses the candidate space at near-flat recall; removing the discriminating identifier rule re-inflates it.}
  \label{tab:rule-ablation}
  \footnotesize
  \setlength{\tabcolsep}{5pt}
  \renewcommand{\arraystretch}{1.15}
  \begin{tabular}{l l r r r}
    \toprule
    \textbf{Make} ($|E_t|$) & \textbf{Hard-rule config} & \textbf{Recall} & \textbf{Mean $|\mathcal{H}|$} & \textbf{Sel.} \\
    \midrule
    \multirow{3}{*}{Make E ($373$)}
      & none                          & 100.0 & 373.0 & 0.0 \\
      & selected (model $+$ variant)  & 86.0  & 50.2  & \textbf{86.5} \\
      & \quad $-$ internal-model      & 93.4  & 364.0 & 2.4 \\
    \addlinespace
    \multirow{3}{*}{Make A ($11{,}934$)}
      & none                              & 100.0 & 11934.0 & 0.0 \\
      & selected (model $+$ type-group)   & 96.4  & 664.2   & \textbf{94.4} \\
      & \quad $-$ model-type-group        & 96.4  & 5479.3  & 54.1 \\
    \bottomrule
  \end{tabular}
\end{table}

\begin{table}[h]
  \centering
  \caption{Expert effort under cross-dataset mapping: one domain expert per make, median values across the deployed makes (a between-make field comparison with comparable tasks, not a within-subject trial). TTC: time to completion; MpH: mappings per hour; Interactions: cases where the available data is insufficient and the expert must seek extra information to complete a mapping; Constraints: hard$+$soft rules authored (CGM only; the baselines use a different rule mechanism, hence ``--''). Best per metric in bold.}
  \label{tab:expert-effort}
  \renewcommand{\arraystretch}{1.15}
  \begin{tabular}{l rrr l}
    \toprule
    \textbf{Setup} & \textbf{TTC (days)} & \textbf{MpH} & \textbf{Interactions} & \textbf{Constraints} \\
    \midrule
    Excel (manual)   & 10.5 & 118 & --  & -- \\
    Rules            & 6.0  & 207 & 112 & -- \\
    Rules + LLM      & 4.5  & 276 & 48  & -- \\
    CGM (Expert) & \textbf{1.5} & \textbf{827} & \textbf{31} & 2 hard + 1 soft \\
    \bottomrule
  \end{tabular}
\end{table}

\bibliography{references}
\end{document}